\documentclass[12pt]{extarticle} 
\usepackage[top=1in, bottom=1in,left=1in,right=1in]{geometry}

\usepackage{algorithm,algpseudocode}

\usepackage{makeidx,amsfonts, amsthm, amssymb, amsmath, graphicx, bbm, color, wrapfig, caption, subcaption, listings, float, mathtools, url, dsfont}
\usepackage{hyperref}

\usepackage[capitalize,nameinlink]{cleveref}

\allowdisplaybreaks

\usepackage[shortlabels]{enumitem}
\setlist{leftmargin=*,itemsep=0.25\itemsep,parsep=0.35\parsep,topsep=0.25\topsep,partopsep=0.29\partopsep}

\newtheorem{theorem}{Theorem}

\newtheorem*{remark*}{Remarks}
\newtheorem{conjecture}[theorem]{Conjecture}

\renewenvironment{proof}[1][\proofname]{ { \it\bfseries #1: }}{\qed}

\newcommand{\cN}{{\mathcal{N}}}

\newcommand{\field}[1]{\mathbb{#1}}
\newcommand{\R}{\field{R}}

\newcommand{\p}{\field{P}}

\DeclareMathOperator{\cov}{Cov}

\DeclareMathOperator{\E}{\mathbb{E}}

\DeclareMathOperator*{\argmin}{argmin}
\DeclareMathOperator*{\eig}{eig}

\newcommand{\mat}[2][rrrrrrrrrrrrrrrrrrrrrrrrrrrrrrrr]{\left[ \begin{array}{#1} #2 \\ \end{array}\right]}

\numberwithin{equation}{section}

\providecommand{\keywords}[1]
{
  \small	
  \textbf{\textit{Keywords: }} #1
}

\title{Recover the spectrum of covariance matrix: \\  a non-asymptotic iterative method}
\author{Juntao Duan\footnote{School of Mathematics, Georgia Institute of Technology, 
    \href{mailto:juntaoduan@gmail.com}{juntaoduan@gmail.com}}
    ,\quad  
    Ionel Popescu\footnote{ University of Bucharest, Faculty of Mathematics and Computer Science, Institute of Mathematics of the Romanian Academy, 
    }
    ,\quad  
    Heinrich Matzinger\footnote{School of Mathematics, Georgia
 Institute of Technology, 
 }
    }

\begin{document}

\maketitle

\begin{abstract}
    It is well known the sample covariance has a consistent bias in the spectrum, for example spectrum of Wishart matrix follows the Marchenko-Pastur law. We in this work introduce an iterative algorithm `Concent' that actively eliminate this bias and recover the true spectrum for small and moderate dimensions.
\end{abstract}

\section{Introduction}
In life when we observe some object, we often collect information from multiple perspectives. For example, we measure differences and similarities of certain animal species by color, sex, height, weight etc, which we call features (or explanatory, independent variables). This unavoidably will end up with a vector collecting those (say $p$) features $(X_1, X_2, \cdots, X_p)$. As we collect sample instances with those $p$ features, we will find each feature $X_i$ follows some probability distribution. For example the animal species' biological sex  follows a Bernoulli distribution with parameter $p\approx 0.5$.

It is fundamental to understand the relations between the $p$ features. In principle, we know everything if we know the joint distribution of the $p$ features, for example cumulative distribution \[
F_{X_1, X_2, \cdots, X_p}(t_1, t_2, \cdots, t_p) = \p(X_1 \le t_1, X_2\le t_2, \cdots, X_p\le t_p)
\]
However, this is unlikely to happen in real world. And estimation of the joint distribution become impossible as $p$ increases due to curse of dimensionality. Simply put, the number of samples required to estimate the joint distribution will grow at $k^p$ where $k$ is the number of samples required to estimate any marginal $X_i$ within requested accuracy.

Since moments (mean, variance, correlation) determines many useful information of random variables, the most natural simplification of the problem is to estimate the moments of those features. Namely,
\[
\E X_1^{a_1} X_2^{a_2} \cdots X_p^{a_p}, \quad a_1, \cdots a_p \ge 0
\]
In particular, collecting first order moments will obtain mean vector $\mu$, and  second order centered moments will be covariance matrix $\Sigma$.
\[
\vec{\mu} = \mat[c]{\E X_1 \\ \vdots \\ \E X_p}, \quad \Sigma = \mat[ccc]{\cov X_1 & \cdots & \cov(X_1, X_p) \\ & \ddots & \\ \cov(X_p, X_1) &\cdots &\cov X_p}
\]
Estimating each element in $\mu$ and $\Sigma$ is not difficult. Since moments can be estimated by taking average statistics from samples. Let $k$-th samples of $X_i$ be $X_i^{\omega_k}$.
\[
\mu_i:=\E X_i \approx \hat{\mu}_i:=\frac{1}{n} \sum_{k=1}^n X_{i}^{\omega_k}, \quad \Sigma_{i,j}:= \cov(X_i,X_j) \approx \hat{\Sigma}_{i,j}:= \frac{1}{n}\sum_{k=1}^n X_{i}^{\omega_k} X_{j}^{\omega_k} - \hat{\mu}_i \hat{\mu}_j
\]
Then for each entry as a random variable,  with the law of large number we can conclude the error goes to zero in the limit and central limit theorem will control the fluctuation of the error is of order $O(\frac{1}{\sqrt{n}})$ where $n$ is the sample size. 

Let us first fix some notations.  Let $X$ be the data matrix with $n$ samples (rows) and $p$ columns (features). Then the sample covariance matrix is 
\[
\frac{1}{n} X^T X -  \frac{1}{n^2} X^T \mathbbm{1} (X^T \mathbbm{1})^T, \quad \mathbbm{1}= \mat[ccc]{1 & \cdots & 1}^T
\]
For simplification of the analysis, we will assume all random variables have mean 0, so that $\vec{\mu}=0$. Then the sample covariance matrix is simplified as 
\[
\hat{\Sigma} = \frac{1}{n} X^T X 
\]
However, as a high dimensional vector or matrix, entry-wise behavior is usually misleading. The matrix $\hat{\Sigma}$ usually exhibits fundamentally different behavior even entries behave as expected. Specifically, the spectrum (eigenvalues) of $\hat{\Sigma}$  has a consistent bias compared with $\Sigma$. For example if we fix $p/n \to c$, spectrum of sample covariance matrix for $X$ with i.i.d. entries mean 0 and variance 1,
 converges to the Marchenko-Pastur law (see \cite{pastur1967}) as $n\to \infty$. 

\begin{figure}[H] 
	\centering
	   \includegraphics[width=1\linewidth]{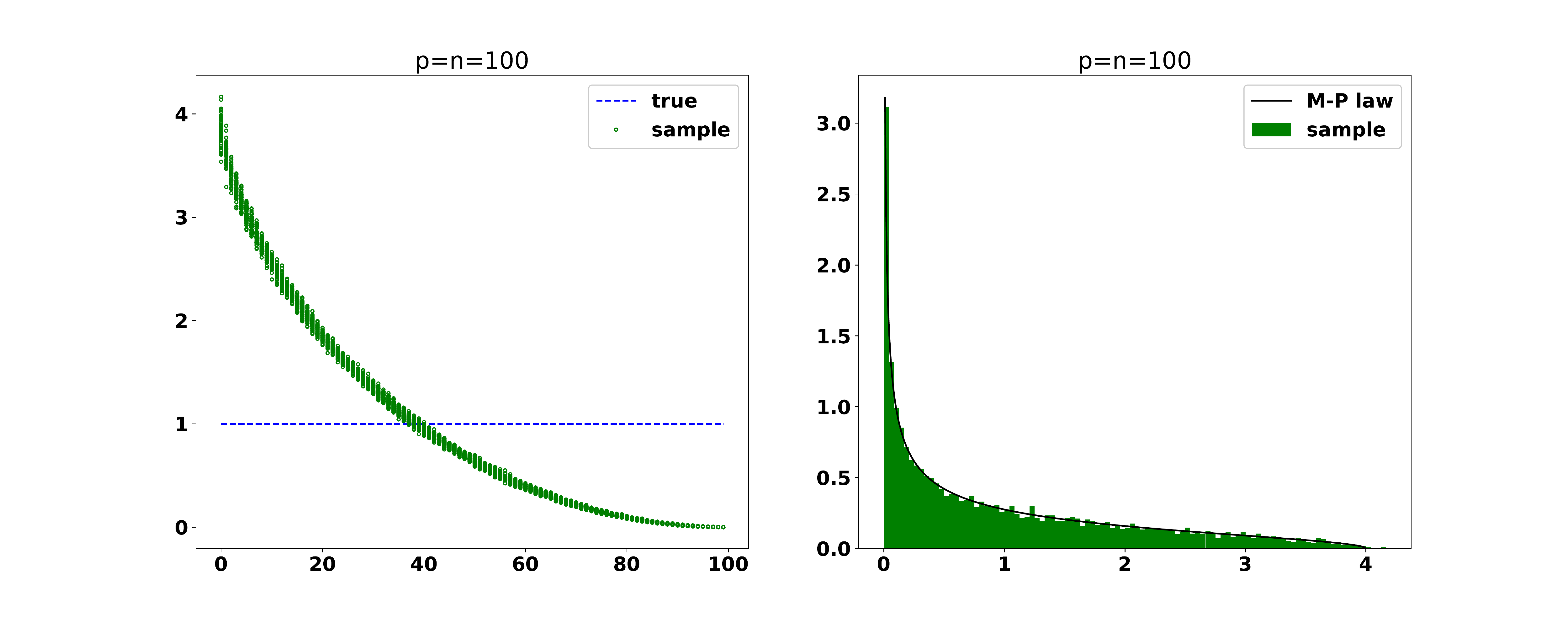} 
	\caption{We take $n=p=100$. The sample covariance matrix ($\hat{\Sigma}$) of standard Gaussian of dimension 100. The true spectrum is $\lambda=1$ since $\Sigma=I$. On the left we see sample spectrum is concentrated around a biased curve. On the right the histogram of the sample spectrum can be fitted to Marchenko-Pastur distribution closely.}\label{fig: sample spectrum Wishart 100}
\end{figure}

In the example as shown in Figure \ref{fig: sample spectrum Wishart 100}, the true covariance matrix is identity, $\Sigma = I$. Since largest eigenvalue of $\hat{\Sigma}$ is very close to $4$, we see largest eigenvalues of the error matrix $\lambda_{max}(\hat{\Sigma} -\Sigma) \approx 4-1=3 $. Therefore $\|\hat{\Sigma} -\Sigma \|\approx 3 $. Similarly, for arbitrary true covariance $\Sigma$, the spectral norm of the error  matrix $\|\hat{\Sigma} -\Sigma \|$ behave similar to a constant (depend on $\frac{p}{n}$) multiple of $\|\Sigma\|$ (see \cite{Kolt2}). Any attempt using sample covariance in a matrix fashion will yield a significant error, for example principle component analysis, MANOVA, factor analysis and linear discriminant analysis etc. in multivariate analysis. 

In many practical applications, the spectrum of the true covariance matrix contains essential information about the structure of the data at hand. Therefore, recovering the true spectrum is critical to understand the behavior of various models we use for the data. In the case  $\Sigma =I$ as shown in figure \ref{fig: sample spectrum Wishart 100}, we can expect a reverting process that may recover the true spectrum from Marchenko-Pastur distribution. In the case of general  covariance matrix $\Sigma$, a series of results (see Silverstein \cite{silverstein1995}, Bai and Yin \cite{bai1988}, Yin, Bai and Krishnaiah \cite{yin1983} ) have shown the spectrum of the sample covariance $\hat{\Sigma}$ converge to the free product of spectrum of $\Sigma$ with a Marchenko-Pastur distribution, provided the spectrum of the true covariance $\Sigma$ converges. The main result is summarized as follows.

\begin{theorem}
	Assume the following. 
	\begin{enumerate}
		\item{} The entries of $X_p=(X_{i,j})_{n\times p}$ are 
			i.i.d. real random variables for all $p$. 
		
		\item{} $E[X_{1,1}]=0$, $E[|X_{1,1}|^2]=1$. 
		
		\item{} Let $p/n \to c >0$ as $p\to \infty$. 
		
		\item{} Let $\Sigma_p$ $(p\times p)$ be non-negative definite symmetric random matrix with spectrum distribution $F^{\Sigma_p}$ (If $\{\lambda_i\}_{1\le i \le p}$ are the eigenvalues of $\Sigma_p$, then $F^{\Sigma_p}=\sum_{1}^{p} \frac{1}{p} \delta_{\lambda_i}(x)$)  such that $F^{\Sigma_p}$ almost surely converges weakly to $F^{\Sigma}$ on $[0, \infty)$. 
		
		\item{} $X_p$ and ${\Sigma}_p$ are independent.
	\end{enumerate}
	Then the spectrum distribution of $W_p= \frac{1}{n}{\Sigma}_p^{1/2}X_p^T X_p {\Sigma}_p^{1/2}$, denoted as $F^{W_p}$  almost surely converges weakly to $F^W$. $F^W$ is the unique probability measure whose Stieltjes transform $m(z)= \int \frac{d F^W(x)}{x-z}$, $z\in \mathbb{C}^+$ satisfies the equation
	\begin{equation} \label{eqn:Marchenko-Pastur equation}
	-\frac{1}{m}=z- c\int \frac{t }{1+tm} d F^{\Sigma}(t) \quad \forall z \in \mathbb{C}^+
	\end{equation}
	
\end{theorem}

In theory, from the limiting distribution of the estimated covariance matrix, one could retrieve $\Sigma$ using the Stieltjes-transform from free probability. This idea, pioneered by  EI Karoui \cite{elkaroui2008},  attempts to discretize the Marchenko-Pastur equation \ref{eqn:Marchenko-Pastur equation} then estimate spectrum by minimizing the residuals. Recently there is a series of follow-up work attempt to improve the discretization and the optimization for example using different discretization and quantization strategy in \cite{ledoit2015, ledoit2017numerical} and moving from complex plane to real line in \cite{li2013}. 

Another type of approach is based on an explicit formula to relate the moments of true limiting spectrum and moments of sample limiting spectrum distribution see for example \cite{bai2010estimation, Valiant}. This formula approximates the finite dimensional relation with an asymptotic normal error. However, this approach is rather restrictive computationally since it has to solve polynomial equations and invert moments back to distribution. Mostly, it can only deal with the case that true spectrum has only a small number of unique values.

However, using limiting result from random matrix theory does not guarantee good performance in finite dimensional case. Even though the estimators are proven consistent in the limit, the rate of convergence was not well understood and in fact very often to be slow. From Figure \ref{fig: compare eigen and Quest 50, 100}, the random matrix approaches `Quest'  \cite{ledoit2015} and `Moment' from \cite{Valiant} usually falls short. Instead, we introduce an iterative algorithm, `Concent' method (black in Figure \ref{fig: compare eigen and Quest 50, 100}), which solves a random optimization problem based on concentration of sample spectrum.  Moreover it also exploited sample covariance eigenvectors to correct sample spectrum. This method exhibits remarkable reconstruction due to the sub-Gaussian concentration behavior of sample spectrum which become effective even with very small $n$ and $p$.  
\begin{figure}[H]
\centering
	\begin{subfigure}[t]{.45\textwidth}
		\centering
		\includegraphics[width=0.98\linewidth]{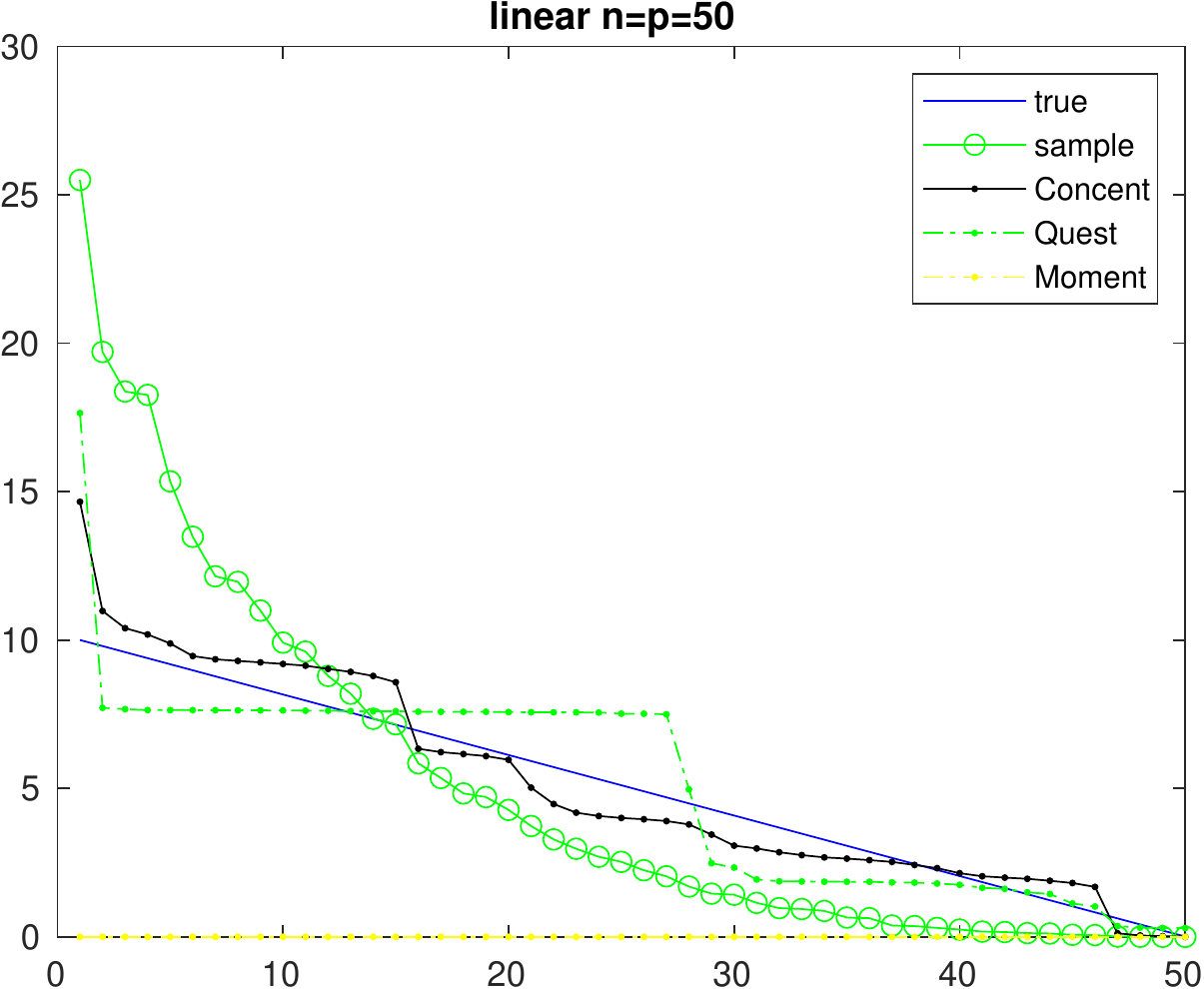}	
	\end{subfigure}
	\begin{subfigure}[t]{.45\textwidth}
		\centering
		\includegraphics[width=.98\linewidth]{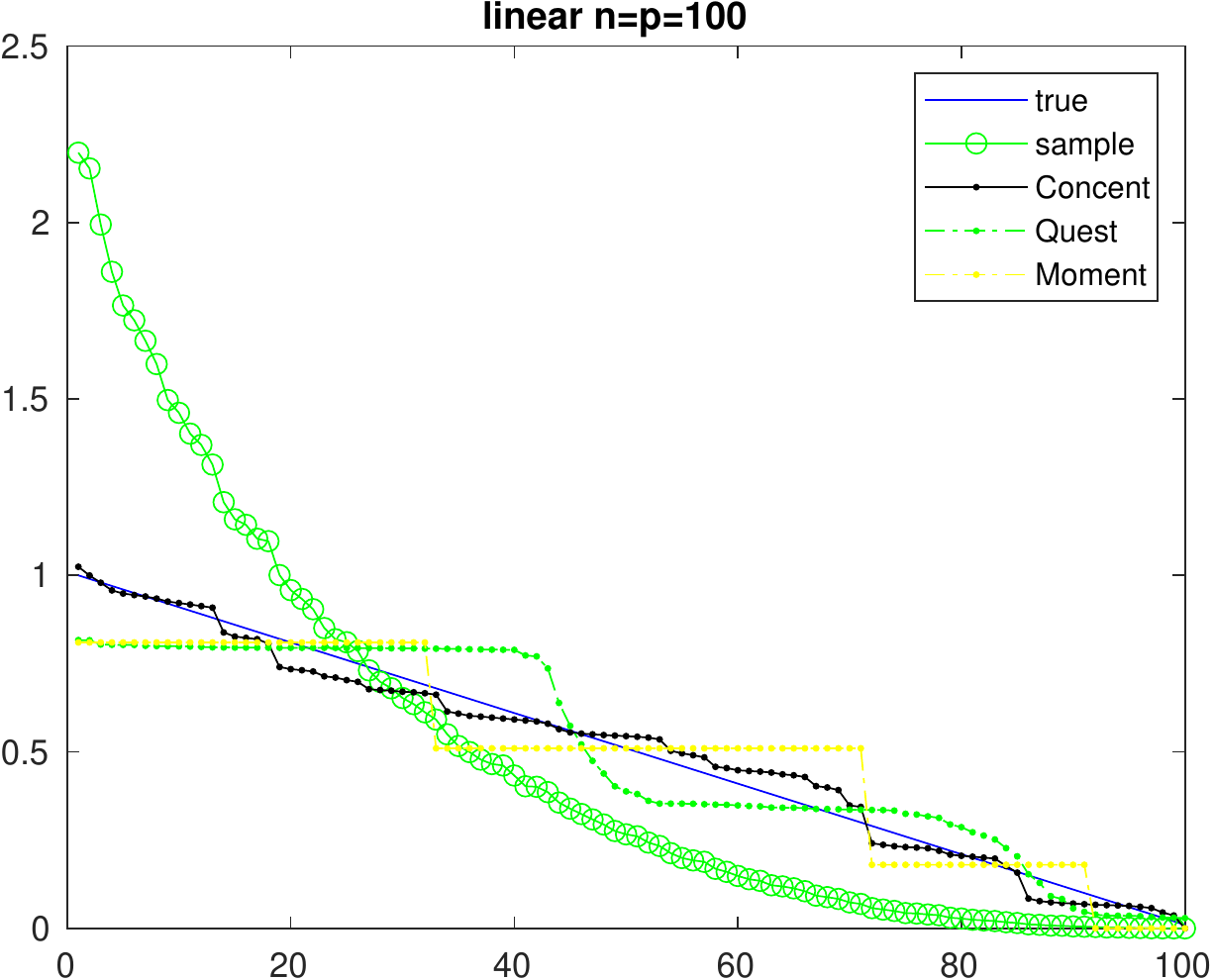}
	\end{subfigure}
	\caption{We take $n=p=50$ on the left, and $n=p=100$ on the right. The true covariance matrix $\Sigma$ has diagonalization $Q\Lambda Q^T$ where the diagonal matrix $\Lambda$ has eigenvalues spaced from 0 to 10 on the left and 0 to 1 on the right uniformly. The sample spectrum (in green) has a convex shape which is significantly different from the true spectrum. The `Quest' estimator form Ledoit \cite{ledoit2015} performs poorly because of its discontinuity from discretization. The `Moment' estimator from \cite{Valiant} is not gaining useful information at all on the left due to true spectrum has values $>1$, which will result higher moments overflow in computer. `Moment' On the right still perform poorly even we restrict spectrum $\le 1$.}\label{fig: compare eigen and Quest 50, 100}
\end{figure}

There is also many other work trying to find better estimator than sample covariance matrix which do not necessarily estimate true spectrum. For example,  under sparsity or low rank condition of the true covariance, shrinkage method on sample covariance exhibits appealing performance, see Stein \cite{Stein}, Bickel and Levina \cite{Bickel} and Donoho \cite{Donoho}.  See \cite{fan2016overview} for a detailed review.

The remaining of this paper is structured as follows. First, we show by various simulations the concentration of the sample spectrum in section \ref{sec:concentration}. This shows the bias in sample spectrum is consistent. Any sample spectrum can be used as a biased baseline. Then in section \ref{sec:random optimization}, we propose an optimization problem and its approximations based on this concentration property. Then in \ref{sec:iterative eigenvector correction} we outline an iterative eigenvector correction algorithm which actively improves approximated optimization solution. At the end, we show some simulations to demonstrate it works well in various settings in section \ref{sec:simulation}.

\section{From concentration of sample spectrum to recovery by optimization} \label{sec:concentration}

\subsection{Concentration of sample spectrum}
Let $\hat{\Lambda}$ (green in Figure \ref{fig: sample spectrum Wishart 100}) be spectrum of sample covariance matrix $\hat{\Sigma}$. The mean of sample spectrum $\E \hat{\Lambda}$  is very far from the true spectrum $\Lambda$ (red in Figure \ref{fig: sample spectrum Wishart 100}).  However $\hat{\Lambda}$ are very concentrated together around $\E \hat{\Lambda}$. It is easily observed but not necessarily easy to prove. We formulate the simplest case (assuming Gaussian) below.
\begin{theorem} \label{thm:concentration of sample spectrum}
	Let $\Lambda$ be the diagonal matrix with the true spectrum, i.e. eigenvalues of $p\times p$ true covariance matrix $\Sigma$ ($\Sigma$ and $\Lambda$ are unknown in practice). Assume all spectrum are sorted decreasingly.  Let $\cN$ be a random $n\times p$ matrix with i.i.d. Gaussian random variables with mean 0 and variance 1, which is unknown in practice as well. Suppose we observe data matrix $X= {\cN}^{T} {\Lambda}^{1/2}$. Then denote $\hat{\Lambda}$ as the spectrum of the sample covariance $W=\frac{1}{n}X^TX = \frac{1}{n}{\Lambda}^{1/2}  {\cN}^{T}\cN {\Lambda}^{1/2}$. Then we have the sample spectrum is concentrated around its mean,
\begin{equation} \label{eqn: concentration of sample}
\p(\|\hat{\Lambda}- \E\hat{\Lambda}\|_{\infty} >\|\Lambda\|_2 t )< Ce^{-cnt^2}	
\end{equation}
\end{theorem}

The proof is based on the Lipschitz continuity of eigenvalue function and a Gaussian concentration inequality, which we present  later. Essentially, this concerns the local statistics for the eigenvalues of finite dimensional sample covariance matrix. For the general case $\cN$ is not Gaussian entries, it is much more complicated and we would not expect a sub-Gaussian tail bound. 

  For the simplest case $\Sigma= I$ (with general random variables) , the universality \cite{tao2012random} would imply the behavior is close to Wishart matrix as long as first four moments are matched for the entries. For the case of $\Sigma \ne I$ (with general random variables), there is no concentration or universality result available. We suspect there is a sub-Gaussian tail if first four moments are matched and sub-exponential tail if not.
  \begin{conjecture}
  Fixing all assumptions in Theorem \ref{thm:concentration of sample spectrum}, except the random matrix $\cN$ have different entries.
  \begin{itemize}
      \item If entries of $\cN$ is not Gaussian but has first four moments matched with Gaussian then we still have sub-Gaussian tail $e^{-nt^2}$.\\
      \item If entries of $\cN$ is not Gaussian and only have first two moments  matched with Gaussian, and first four moments are finite then we  have sub-exponential tail $e^{-nt}$.
  \end{itemize}
  \end{conjecture}

\begin{figure}[H] 
	\centering
	\includegraphics[width=1.05\linewidth]{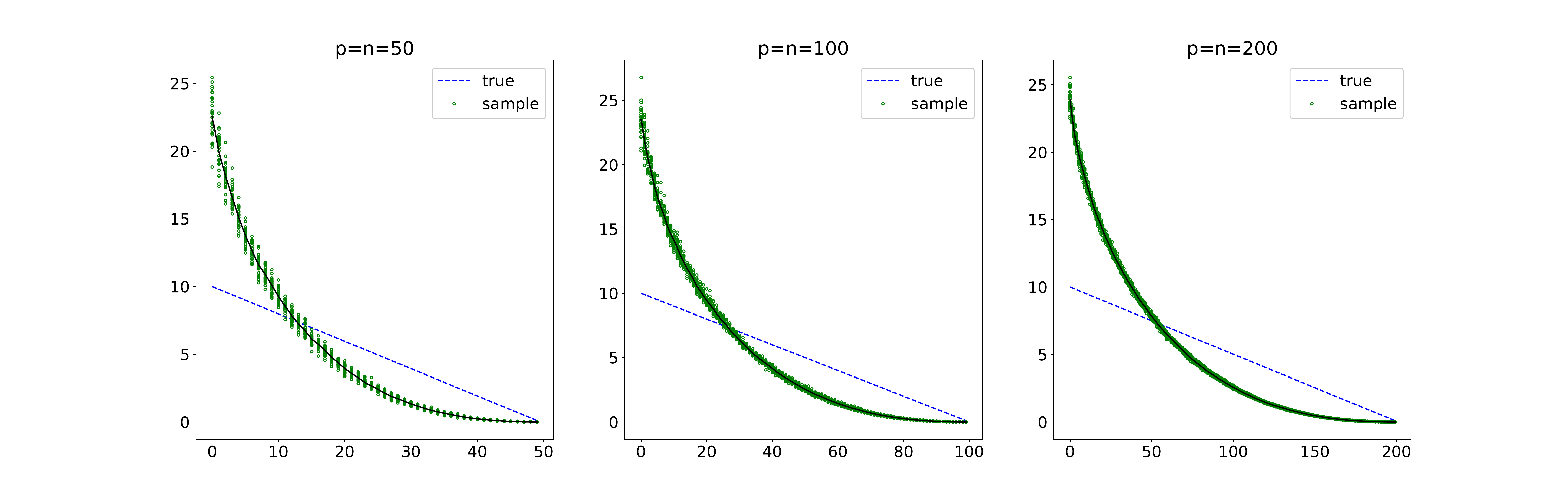}
	\includegraphics[width=1.05\linewidth]{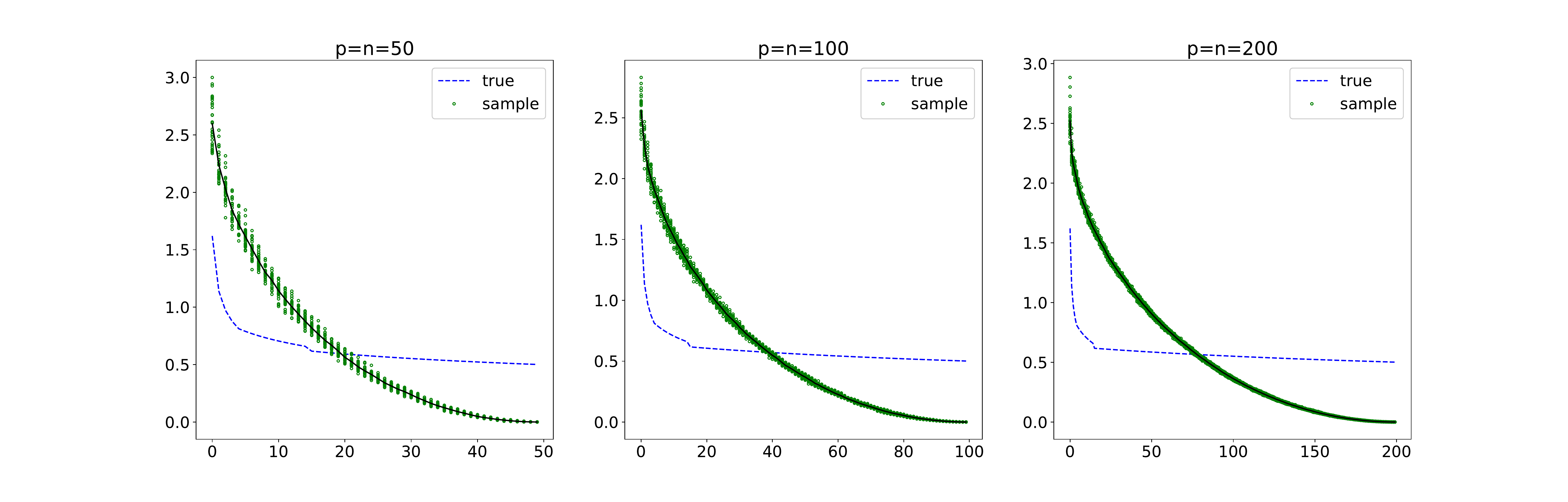}
	\caption{The pictures in the first row has true covariance $\Sigma$ with eigenvalues ranging from 0 to 10 evenly of step size $1/p$. The graph in the second row has true spectrum $\Sigma$ with eigenvalues ranging from 0 to 10 evenly of step size $1/p$.}\label{fig: concentration 2 linear}
\end{figure}

From Figure \ref{fig: concentration 2 linear}, clearly sample spectrum are concentrated around a convex curve which is significantly different from the true spectrum ($\Sigma \ne I$). 
\begin{proof}
The difficulty in proving such concentration result mainly due to the complexity of eigenvalue function.  We will show the eigenvalue function is Lipschitz. Let eigenvalue $diag(\Lambda) = (\lambda_1, \cdots, \lambda_p)$ be the mapping $ f =\R^{n\times p} \to \R^p$ that $f(\cN): Q^{T}\left( \frac{1}{n}{\Lambda}^{1/2}  {\cN}^{T}\cN {\Lambda}^{1/2} \right) Q = \Lambda  $ where $Q$ is orthogonal matrix. 
Then we compute a perturbation 
\begin{align*}
    \| \Lambda - \Lambda'\|_{\infty} 
    & = \| \Lambda - \Lambda'\|_2 \\
    & = \| Q^{T}\left( \frac{1}{n}{\Lambda}^{1/2}  ({\cN}^{T}\cN - {\cN'}^{T}\cN') {\Lambda}^{1/2} \right) Q \|_2 \\
    & \le \frac{1}{n} \|{\Lambda}\|_2  \| {\cN}^{T}\cN - {\cN'}^{T}\cN'\|_2 \\
    & \le \frac{1}{n} \|{\Lambda}\|_2 ( \| {\cN}^{T}(\cN - \cN')\|_2+ \| {(\cN-\cN')}^{T}\cN'\|_2)
\end{align*}
Notice for rectangular matrices $A,B$, $\|AB\|2\le \|A\|_2\|B\|_F$. Then we conclude 
\[
\| {\cN}^{T}(\cN - \cN')\|_2 \le \| {\cN}^{T}\|_2 \|\cN - \cN'\|_F = \| {\cN}\|_2 \|\cN - \cN'\|_F 
\]
and similarly
\[
\| (\cN - \cN')^{T}{\cN'}\|_2 =\| {\cN'}^{T}(\cN - \cN')\|_2 \le \| {\cN'}\|_2 \|\cN - \cN'\|_F 
\]
This leads to the bound on the variation of eigenvalues
\[
\| \Lambda - \Lambda'\|_{\infty} \le \frac{1}{n} \|{\Lambda}\|_2 (\|\cN\|_2 +\|\cN'\|_2) \|\cN - \cN'\|_F
\]
For any $f$ has Lipschitz constant $L \le \frac{1}{n} \|{\Lambda}\|_2 (\|\cN\|_2 +\|\cN'\|_2)$, Of course this is random variables. However, for Gaussian random matrix $\cN$ we have the concentration of the norm (or singular value) \cite{rudelson2010non}
\[
\p( \|\cN\|_2 > C(\sqrt{n}+\sqrt{p}) +t ) \le 2e^{-ct^2}
\]
With overwhelming probability, the function $g =\|\cdot \| \circ f$ has Lipschitz constant $L\le \frac{1}{n}\|{\Lambda}\|_2 C(\sqrt{n}+\sqrt{p})$. Then recall Gaussian concentration inequality (can be found in many textbooks for example \cite{boucheron2013concentration}) states that for $g: \R^N \to \R$ with Lipschitz constant $L$ then for a Gaussian random vector, we have 
\[
\p \left(\|g(X) - \E g(X)\|>t\right)<2 e^{-\frac{t^2}{2L^2}}
\]
Therefore apply with $L=\frac{1}{n}\|{\Lambda}\|_2 C(\sqrt{n}+\sqrt{p}) =\|{\Lambda}\|_2O(1/\sqrt{n})$ (we assumed $p\le c n$), we obtain  bound
\[
\p \left(\|\hat{\Lambda} - \E \hat{\Lambda}\|_{\infty} >\|\Lambda\|_2 t \right)<C e^{-cnt^2}
\]
\end{proof}

\subsection{Random optimization} \label{sec:random optimization}
Let $\Lambda$ be the true spectrum, and $\hat{\Lambda}$ be the sample covariance spectrum. Due to the concentration in previous section, we propose the following optimization,
\begin{equation}\label{eqn: optimization}
\min_{{D}\geq 0} \sum_{k=1}^K \left\| \hat{\Lambda}- \eig({D}^{1/2} \cN_k^T \cN_k {D}^{1/2} )    \right\|
\end{equation}
where $\cN_i$ is $n\times p$  random matrix with i.i.d. standard normal random variables, and $\eig(\cdot)$ is computing the eigenvalues and sort in descending order. One note that the norm could be $\ell_1, \ell_2$ or any vector norm. However, by analyzing the performance on simulation, we did not observe too much difference for various norms.  for computational reason $\ell_2$ is taken in the following discussion.

One caveat of this formulation is the optimization problem does not have convex structure so that it can not be solved by fast algorithms available in convex optimization literature. The reason for the complexity in the objective function is due to $\eig(\cdot)$ need to compute eigenvalues and sort afterwards. One could use a generic global optimizer search engine (e.g. Genetic algorithm)  but it will be extremely slow and essentially not applicable for large dimension. So we replace it with a approximation of the problem. First, we translate the problem into a exact relation if we knew the true spectrum $\Lambda$.
\begin{equation} \label{eqn: optimization_true_ratio}
\Lambda_{concent} := \argmin_{{D}\geq 0} \sum_i \left\| \hat{\Lambda}- R_k{D}    \right\|
\end{equation}
where $R_k$ is a vector obtained by element-wise division below
\[
R_k= \frac{\eig({\Lambda}^{1/2} \cN_k^T \cN_k {\Lambda}^{1/2} )}{\Lambda}
\]

Of course, $R_k$ is not obtainable, so we replace it with an estimator 
\[
\hat{R_k}= \frac{\eig({\tilde{\Lambda}}^{1/2} \cN_k^T \cN_k \tilde{\Lambda}^{1/2} )}{\tilde{\Lambda}}
\]
where $\tilde{\Lambda}$ is any reasonable estimator of the  covariance spectrum. In principle, one can iteratively find a sequence of such  estimators $\tilde{\Lambda}_k$. We use the sample spectrum to start the iteration $\tilde{\Lambda}_0= \hat{\Lambda}$. Thus we arrived at 
\begin{equation}\label{eqn: optimization_ratio}
\Lambda_{concent}= \argmin_{{D}\geq 0} \sum_k \left\| \hat{\Lambda}- \hat{R_k}{D}    \right\|
\end{equation}

Now let's derive an explicit formula for $\ell_2$ minimization. The objective function can be rewritten as 
\[
f=\sum_k \left\| \hat{\Lambda}- \hat{R_k}{D}    \right\|^2= \sum_{k=1}^{K}\sum_{j=1}^{p} (\hat{\lambda}_j-\hat{R}_{kj}d_j)^2
\]
Then set partial derivatives $\partial f/ \partial{d_j}$ to be zero, we found 
\[
d_j=\frac{\hat{\lambda}_j \sum_{k=1}^K \hat{R}_{kj}}{\sum_{k=1}^K \hat{R}_{kj}^2 }
\]
Here  $d_j$ will serve as an estimator of the true spectrum $\lambda_j$. And $\hat{R}_{kj}$'s are approximates of ${\hat{\lambda}_j}/{\lambda_j}$. In principle, the simplest approximation would be taking average of such ratio to get a naive estimator ${\hat{\lambda}_j \sum_{k=1}^K \hat{R}_{kj}}/{K}$.
But instead our $\ell_2$ minimization give a second order correction of this naive approach. However, this approach has many parts replaced by estimators instead of the true, thus we will propose a follow up eigenvector correction procedure.



\subsection{An eigenvector correction } \label{sec:iterative eigenvector correction}
We start with any estimator, say with $\Lambda_0=\Lambda_{concent}$. Then in $k+1$-th step, we simulate a sample covariance matrix and diagonalize it
\[
W_{k}=\Lambda_k^{1/2} \cN^T \cN \Lambda_k^{1/2} /n \quad \to\quad W_k= V_k D_k V_k^T
\]
Then we obtain the next estimator by the diagonal elements of the matrix $V_k \hat{\Lambda} V_k^T$.
\[
\Lambda_{k+1}= diag(V_k \hat{\Lambda} V_k^T)
\]


We give a heuristic argument here to explain its effectiveness. Let $\hat{W}=Q^T\Lambda^{1/2} \cN^T \cN \Lambda^{1/2} Q/n$ be the given sample covariance matrix, then the true covariance matrix is $W= Q \Lambda Q^T$, then diagonal elements of $Q^T \hat{W} Q$ will be a good estimator of true spectrum. Since 
\[
diag(Q^T \hat{W} Q) = q_k^T \hat{W} q_k= \lambda_k \frac{1}{n}\sum_{i=1}^{n} \cN_{i,k}^2 \to \lambda_k
\]
where $\cN_{i,k}$ is $i,k$-th entry of $\cN$. In our procedure, the $V_k$ will play a similar role as $Q$. One limitations about this procedure is that it does not apply to high dimensional setting, for example $p\ge 10^4$. The computation would be too demanding due to the eigen-decomposition used in the iteration. However for small or moderate dimensions (for example $p= 10^3$), the iteration converges fast and usually less than $10$ iterations would be sufficient.

Combining the two approach, we propose the following 'Concent' algorithm for spectrum recovery.
\begin{algorithm}[H]
\caption{`Concent': Eigenvector corrected random optimization }\label{alg:cap}
\begin{algorithmic}
\Require $n, p \geq 0$, data matrix $X$.
\Require Set  $loops \ge 10$, averaging  $K \ge 10$.
\State $\hat{\Lambda} \gets \eig (X^T X/n)$
\State Initialize: $\Lambda_1 \gets \hat{\Lambda}$
\For{$i=1:loops$}
    \begin{itemize}
        \item \; \text{Approximated random optimization}
    \end{itemize}
    \State Generate random $n\times p$ standard normal matrix  $\cN_1,\cdots, \cN_K$,
    \For{$k=1:K$}
        \State  Create ratio vector $\hat{R}_k \gets  diag( \eig({\Lambda_i}^{1/2} \cN_k^T \cN_k {\Lambda_i}^{1/2} ) /{\Lambda_i} )$
    \EndFor
    \State Create diagonal matrix $S = diag (\frac{ \sum_{k=1}^K \hat{R}_{kj}}{\sum_{k=1}^K \hat{R}_{kj}^2 }, \cdots \frac{ \sum_{k=1}^K \hat{R}_{kj}}{\sum_{k=1}^K \hat{R}_{kj}^2 } ) $
    \State $\Lambda_i \gets \hat{\Lambda} S$
    \begin{itemize}
        \item \; \text{Eigenvector correction}
    \end{itemize}
    \State Generate  normal matrix  $\cN$, and compute $W = \Lambda_i^{1/2} \cN^T \cN \Lambda_i^{1/2} /n $
    \State Diagonalize $W= V D V^T$
    \State $\Lambda_{i+1} \gets diag(V \hat{\Lambda} V^T)$
\EndFor
\end{algorithmic}
\end{algorithm}

\section{Simulation}\label{sec:simulation}
Here we show the simulations on various settings of our method. We also compare with `Quest' estimator form Ledoit \cite{ledoit2015} and `Moment' estimator from \cite{Valiant}.

\subsection{Simulated spectrum}
We have shown our 'Concent' method performing well for linear spectrum in Figure \ref{fig: compare eigen and Quest 50, 100}.  We next exam the case that true spectrum has a convex or concave shape.
\begin{figure}[H] 
	\centering
	\begin{subfigure}[t]{.45\textwidth}
		\centering
		\includegraphics[width=.99\linewidth]{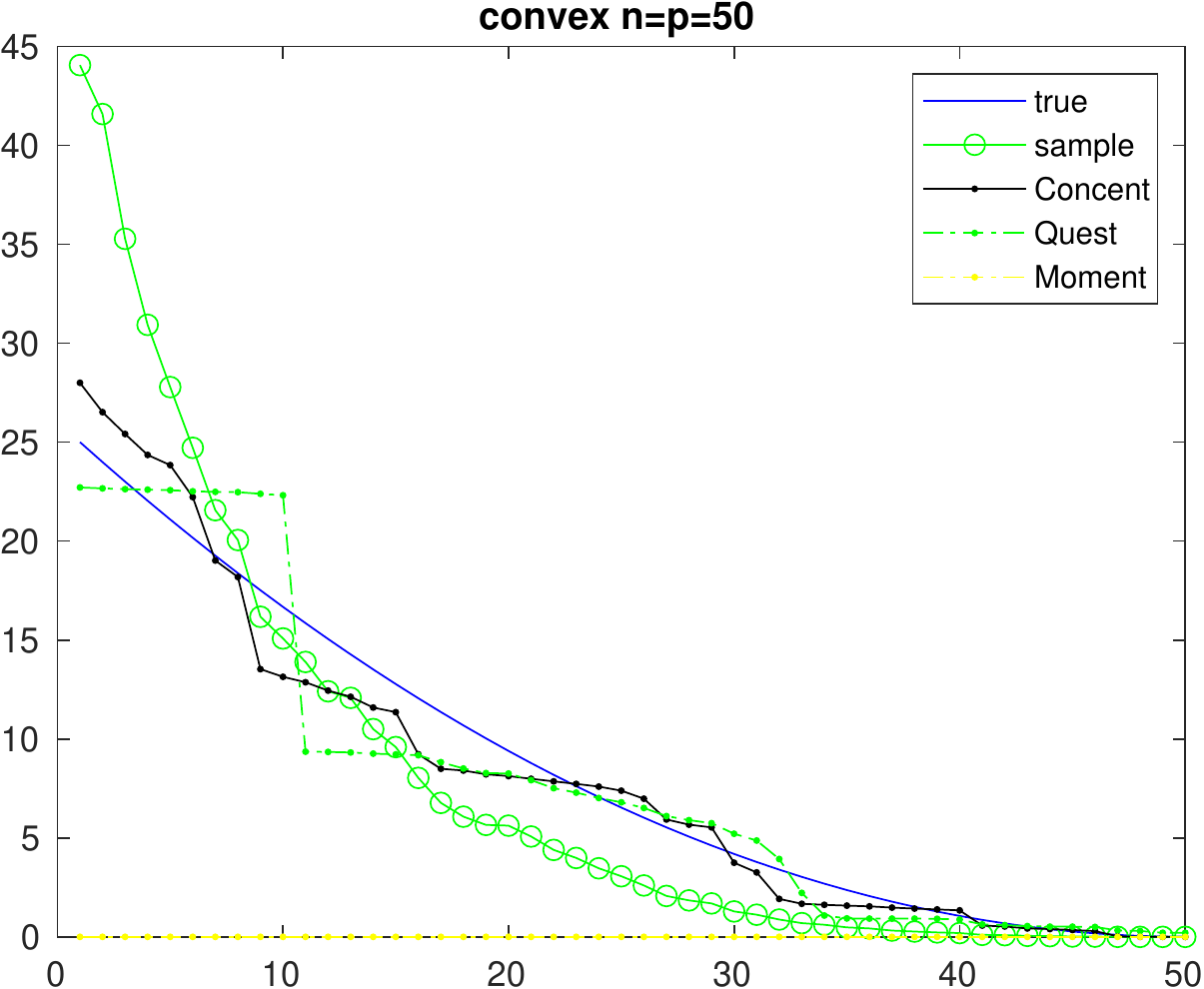}
	\end{subfigure}
	\begin{subfigure}[t]{.45\textwidth}
		\centering
		\includegraphics[width=.98\linewidth]{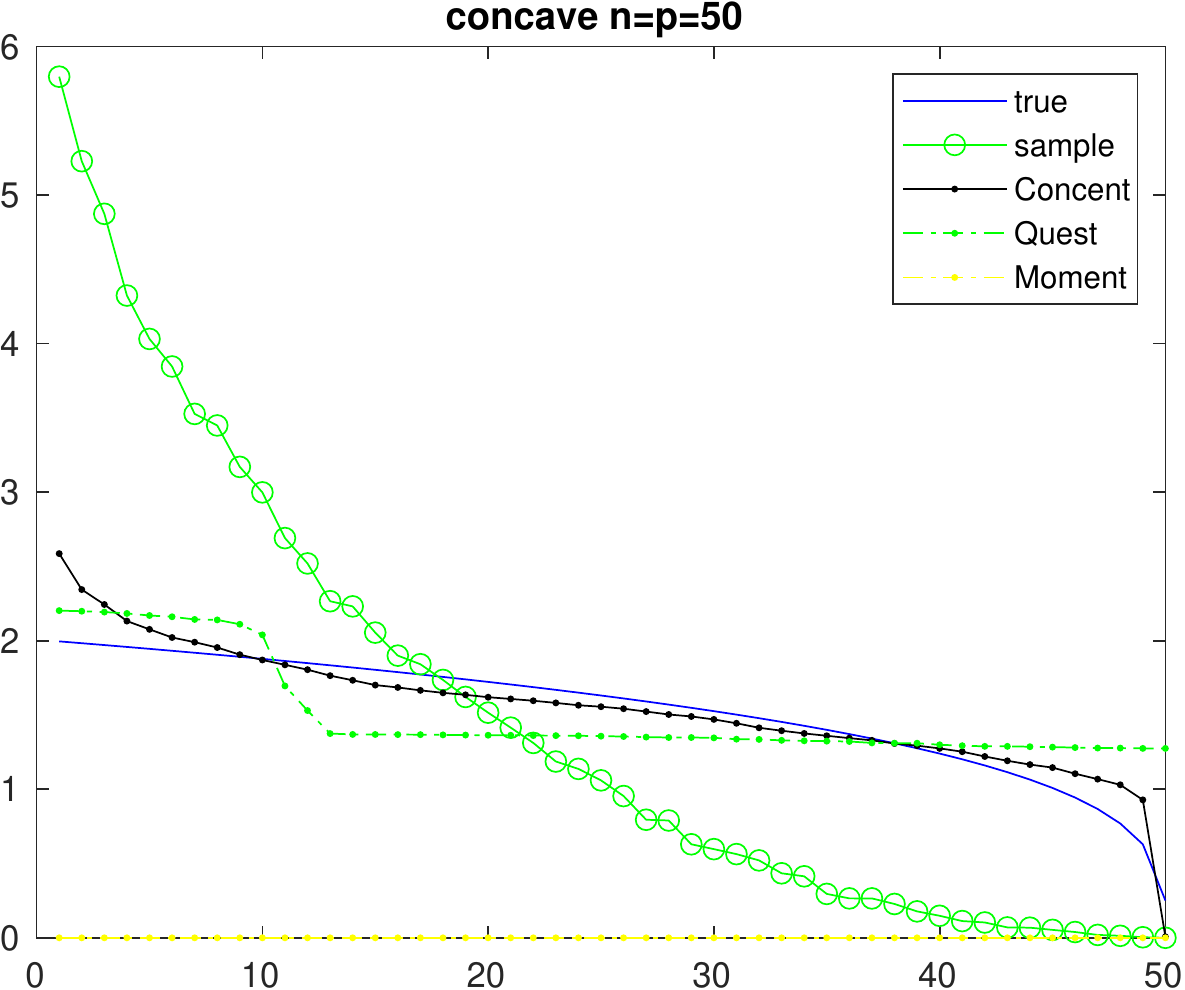}
	\end{subfigure}	
	\caption{ The true spectrum on the left takes $x^2$ where $x$ is evenly spaced in $(0,5]$. On the right the true spectrum is taking $x^{0.3}$.}\label{fig:convex and concave spectrum 50}
\end{figure}

When we deal with spectrum of special unknown structure, it's still possible to recover the smoothing approximation of the true spectrum using our `Concent' algorithm. Here is a simulation with spectrum of step shape and sparse shape.
\begin{figure}[H] 
	\centering
	\begin{subfigure}[t]{.45\textwidth}
		\centering
		\includegraphics[width=.98\linewidth]{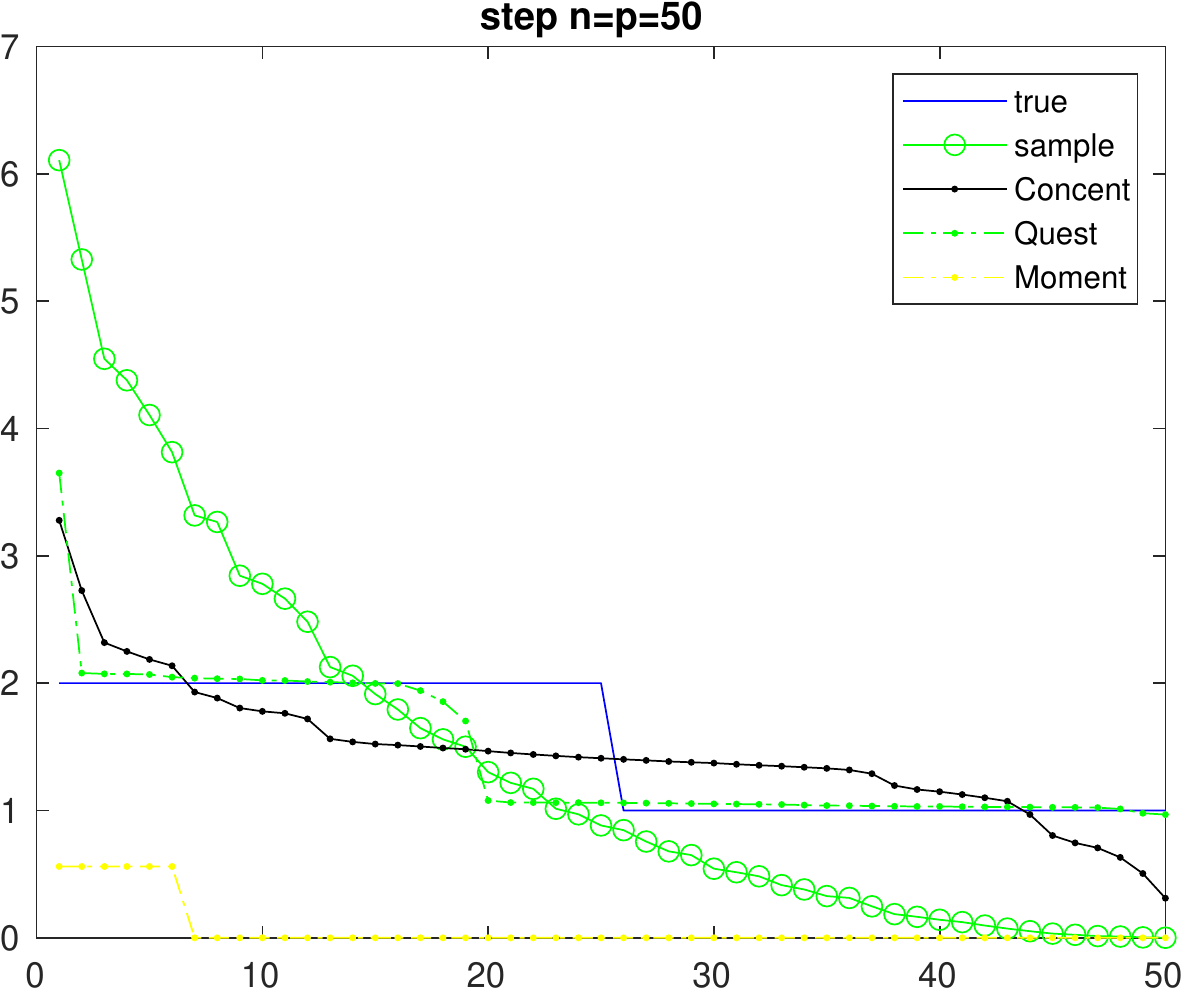}
	\end{subfigure}
	\begin{subfigure}[t]{.45\textwidth}
		\centering
		\includegraphics[width=.99\linewidth]{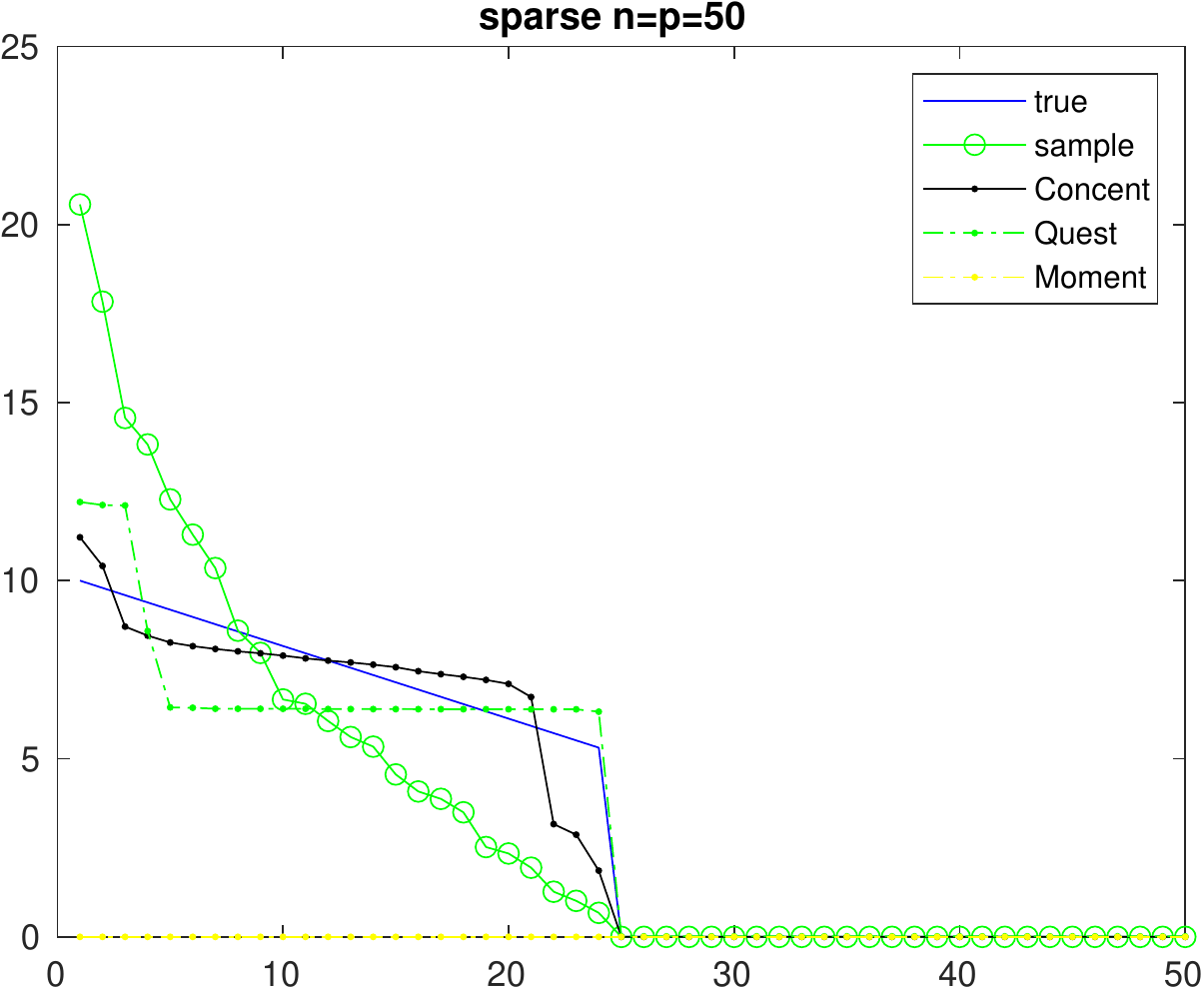}
	\end{subfigure}	
	\caption{On the left, the true spectrum are half 2's and half 1's. On the right the true spectrum is taking by zero out last half of the linear spectrum.}\label{fig:structured spectrum}
\end{figure}

\subsection{Real world data}
We compare the result with the true spectrum generated from large sample size real stock data. The `true' spectrum is taken from  50 stocks with 1000 days. $n/p = 20$ which means the `true' spectrum is relatively close to the spectrum of real stock covariance matrix.
 
\begin{figure}[H] 
\centering
	\begin{subfigure}[t]{.45\textwidth}
		\centering
		\includegraphics[width=.99\linewidth]{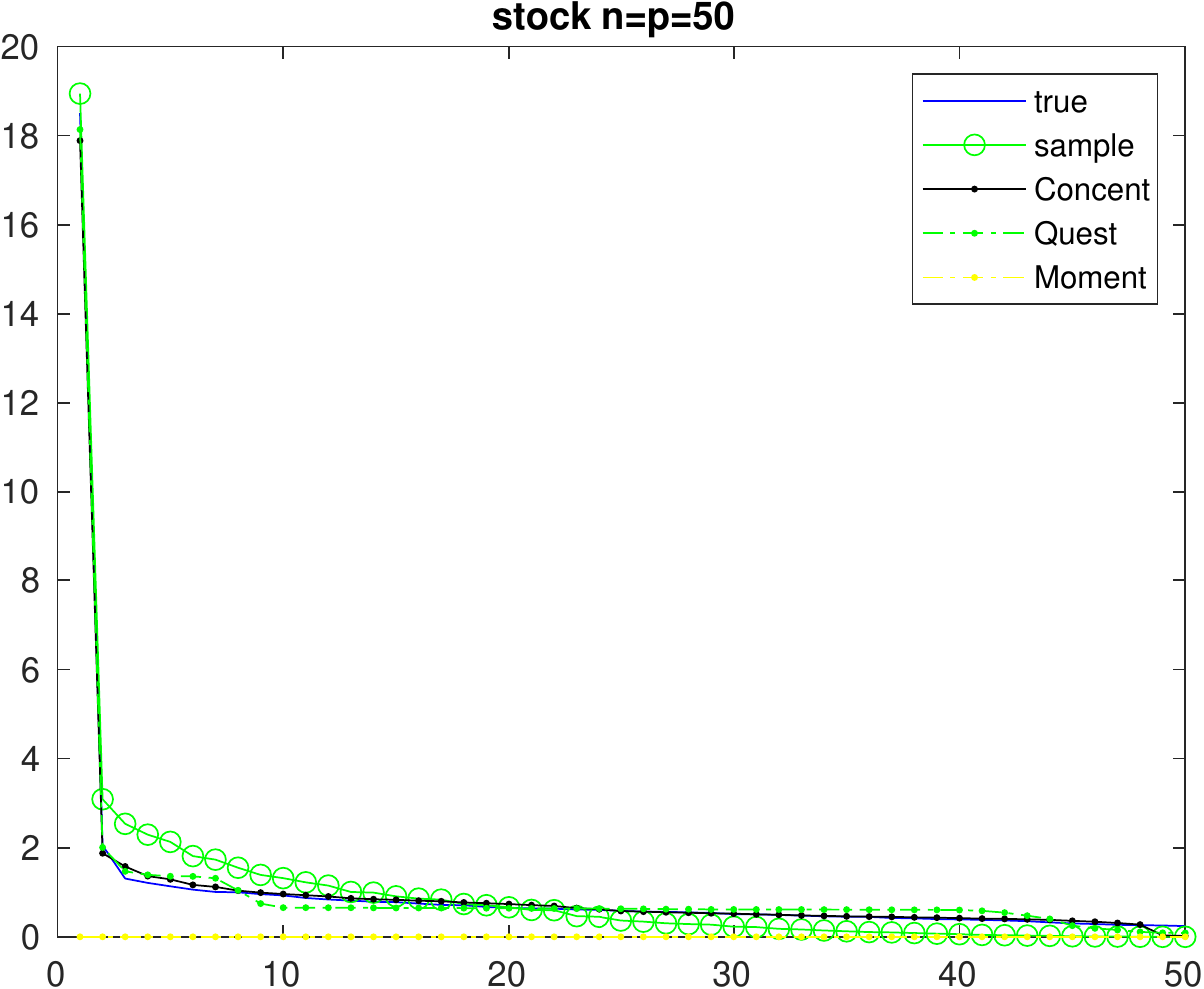}
	\end{subfigure}
	\begin{subfigure}[t]{.45\textwidth}
		\centering
		\includegraphics[width=.99\linewidth]{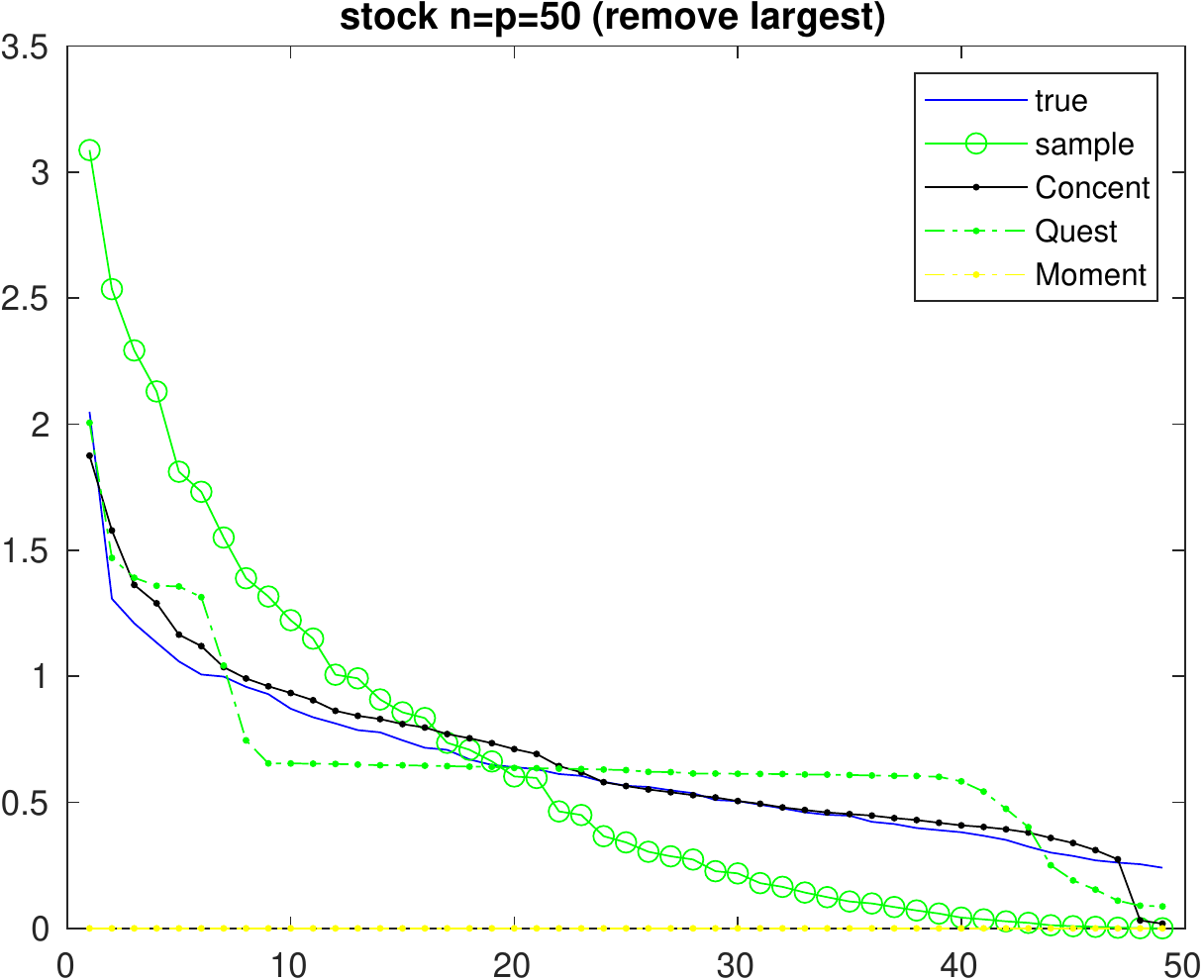}
	\end{subfigure}
	\caption{On the right, we removed the largest eigenvalue to make it easy to see the difference.}
\end{figure}

Another example we study is amazon reviews dataset. We take 50 products with 4082 reviews for each. Therefore $n/p \approx 80$, we are confident the sample spectrum from this data is close to the true spectrum.
\begin{figure}[H] 
\centering
	\begin{subfigure}[t]{.45\textwidth}
		\centering
		\includegraphics[width=.95\linewidth]{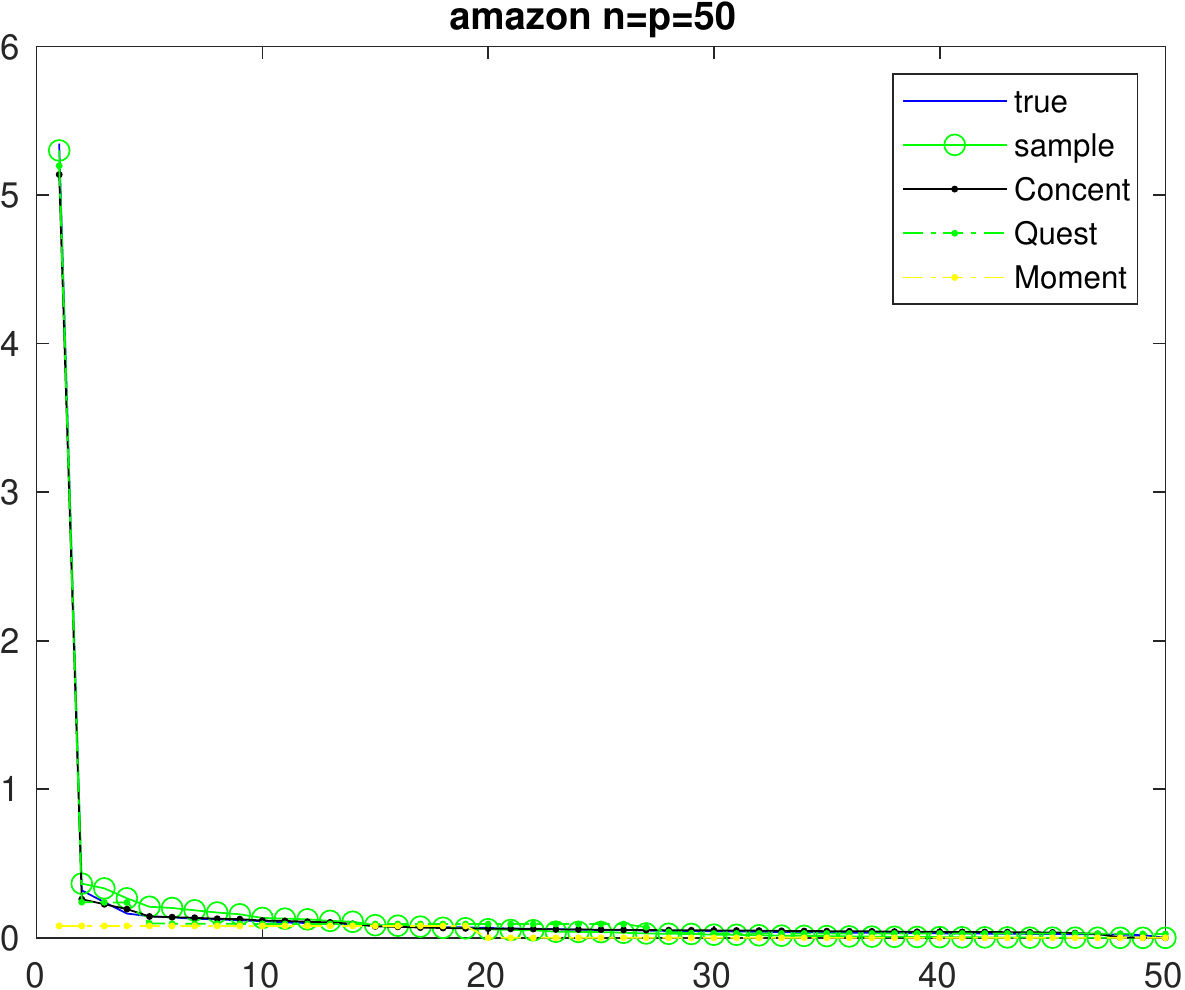}
	\end{subfigure}
	\begin{subfigure}[t]{.45\textwidth}
		\centering
		\includegraphics[width=.99\linewidth]{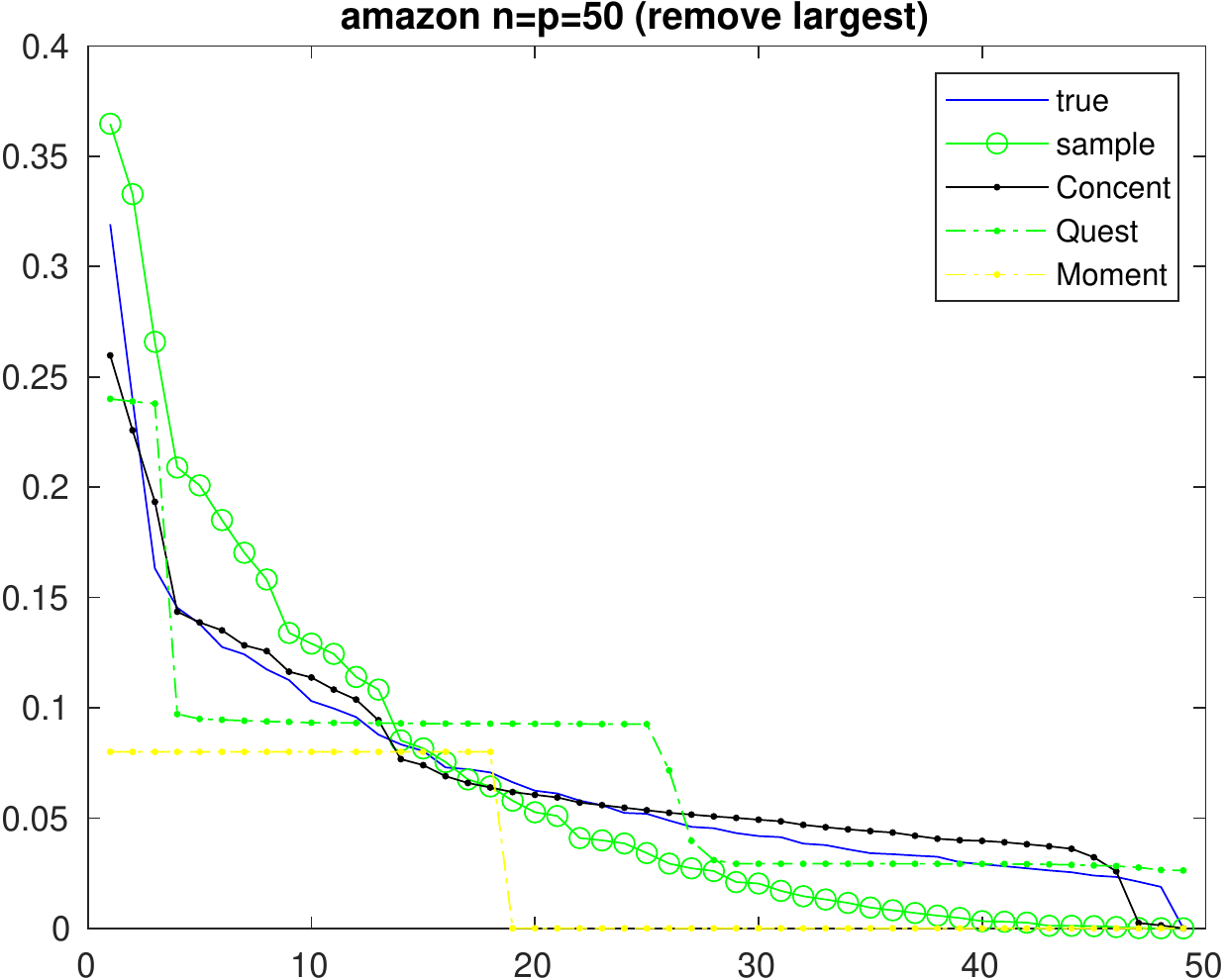}
	\end{subfigure}
	\caption{On the right, we removed the largest eigenvalue to show the difference in bulk eigenvalues.}
\end{figure}

\begin{figure}[H] 
    \centering
	\includegraphics[width=.45\linewidth]{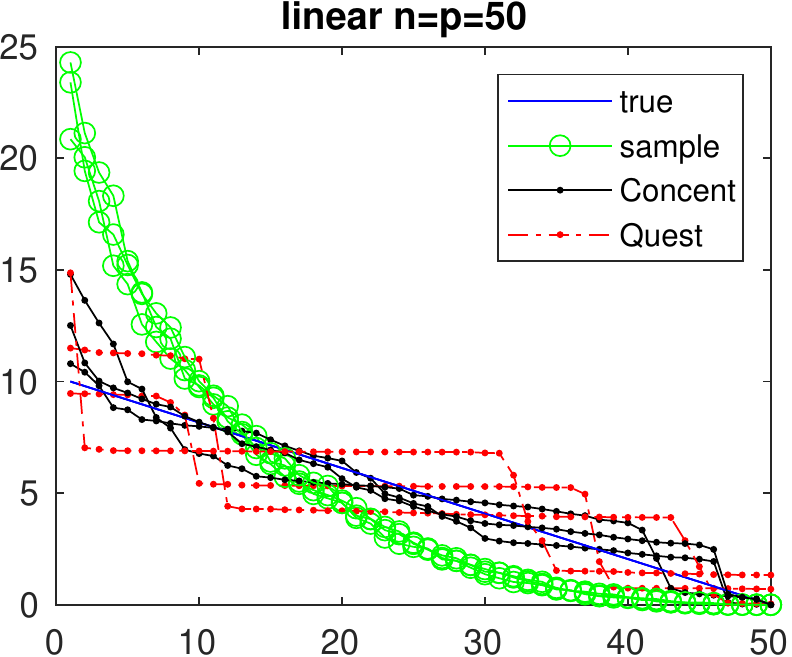}
	\caption{We put 3 simulated recovery together. Even the sample spectrum is very concentrated but the Quest varies significantly.}\label{fig:Quest weak}
\end{figure}

In the majority of those cases, `Concent'  outperforms others. There is another significant advantage of `Concent'. That is its robustness against to the random generated samples. In other words, taken any sample spectrum, `Concent' would be able to use it to recover the true spectrum due to powerful finite dimensional concentration of sample spectrum. On the other hand, in Figure \ref{fig:Quest weak}, `Quest' (in red) varies quite significantly even when sample spectrum (in green) changes very little.

Moments method generally does not produce relevant result.  The total number of moments can be computed are too few (around 10). When eigenvalues are large than 1, large moment will blow up. In the case of eigenvalues are less than 1, then higher order moments will be close to zero and produce no information. 


\section{Conclusion}
  We derive a concentration of sample spectrum result and propose a random optimization to recover the true spectrum. Our method of recovering the spectrum is based on finite dimensional concentration of measure behavior so that it provides a competitive performance for small and moderate dimensional covariance matrix. It is much more stable compared with `Quest' and `Moment' method  which are based on properties of the limiting random matrix behavior. From simulations we showed our algorithms overcome several weakness of `Quest' and `Moment' method. `Quest' method is very sensitive to small changes in sample spectrum and usually produce a discontinuous estimator. `Moment' method does not work properly for small or moderate dimensions and will blow up for eigenvalues larger than 1. 

There are several limitations of our method. First, it has expensive diagonalization procedure which will be hard to implement for large dimensions. Second, for discontinuous true spectrum, the recovery is only possible if the structure is known otherwise it produces smoothing approximations of the true spectrum.

\bibliography{SpectrumRecovery}
\bibliographystyle{plain}
\end{document}